# Localization & Mapping Requirements for Level 2+ Autonomous Vehicles

Tyler G.R. Reid, Andrew Neish, & Brian Manning
*Xona Space Systems*

**ABSTRACT**

Autonomous vehicles are being deployed with a spectrum of capability, extending from driver assistance features for the highway in personal vehicles (SAE Level 2+) to fully autonomous fleet ride sharing services operating in complex city environments (SAE Level 4+). This spectrum of autonomy often operates in different physical environments with different degrees of assumed driver in-the-loop oversight and hence have very different system and subsystem requirements. At the heart of SAE Level 2 to 5 systems is localization and mapping, which ranges from road determination for feature geofencing or high-level routing, through lane determination for advanced driver assistance, to where-in-lane positioning for full vehicle control. We assess localization and mapping requirements for different levels of autonomy and supported features. This work provides a framework for system decomposition, including the level of redundancy needed to achieve the target level of safety. We examine several representative autonomous and assistance features and make recommendations on positioning requirements as well map georeferencing and information integrity.

**INTRODUCTION**

In 2019, there were 33,244 fatal car crashes in the United States, resulting in 36,096 fatalities (National Highway Traffic Safety Administration, 2022). This represents 94% of all transportation related deaths and is 1.11 fatalities per 100 million vehicle miles (Bureau of Transportation Statistics, 2022; National Highway Traffic Safety Administration, 2022). A major push for the deployment of Advanced Driver Assistance Systems (ADAS) is to improve vehicle safety, where it has been argued that the current limiting factor is the human driver (Reid, Houts, et al., 2019). ADAS is deployed today in various forms ranging from active safety features such as automatic emergency breaking to hands-off self-driving features on controlled access divided highways, commonly known as Level 2 or Level 2+ autonomy. As semi-autonomous systems evolve so does the respective roles of the human and virtual drivers.

Here we examine the localization and mapping requirements for Level 2+ systems. Figure 1 shows a simplified view of autonomous driving as the following steps: (1) Localization gives the first order context, the location of roads and infrastructure; (2) Perception detects pedestrians, cyclists, and other vehicles; (3) Prediction anticipates where the objects will be going in the relevant time horizon; (4) Planning your path based on the rules of the road and the others we share the road with; and (5) Control implements throttle, braking, and steering to achieve the planned path. In this process, localization is the foundation, forming the first step in vehicle situational awareness. Errors in localization cascade downstream, resulting in localization having some of the most stringent requirements (Reid, Houts, et al., 2019).

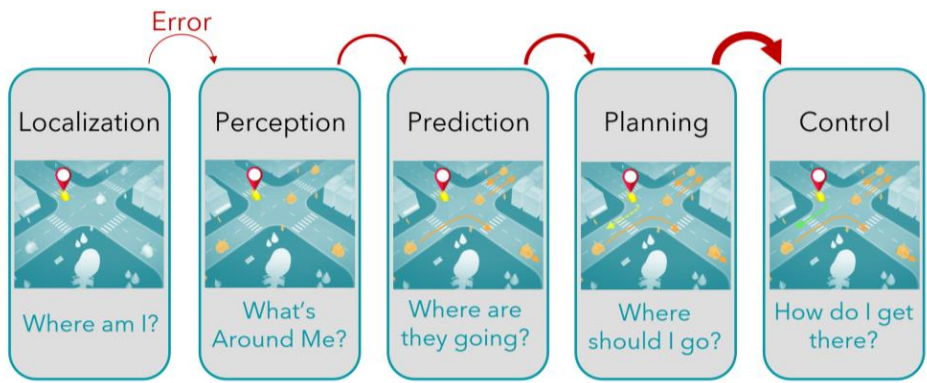

*Figure 1: Localization as the foundation in autonomous driving. Errors in localization can cascade to other systems downstream, leading to localization having the among the strictest requirements.*

Localization requirements are driven by two components, geometry and safety as summarized in Figure 2. Geometry is dictated by the width and curvature of the road, size of the vehicle, and desired level of situational awareness for the intended function. Situational awareness in this context is broken down into three categories: (1) which road, (2) which lane, and (3) where in the lane. Each is important in unlocking new features in autonomy and each comes with new and more stringent requirements. The other factor is safety. The Target Level of Safety (TLS) drives the statistics, or how well location needs be known in order to close the overall safety case. Both geometry and safety also contribute to the semantic and geometry information integrity requirements contained in the map. This builds on previous work in deriving the localization requirements for Level 4 full self-driving vehicles (Reid, Houts, et al., 2019) and extending this to Level 2+ systems.

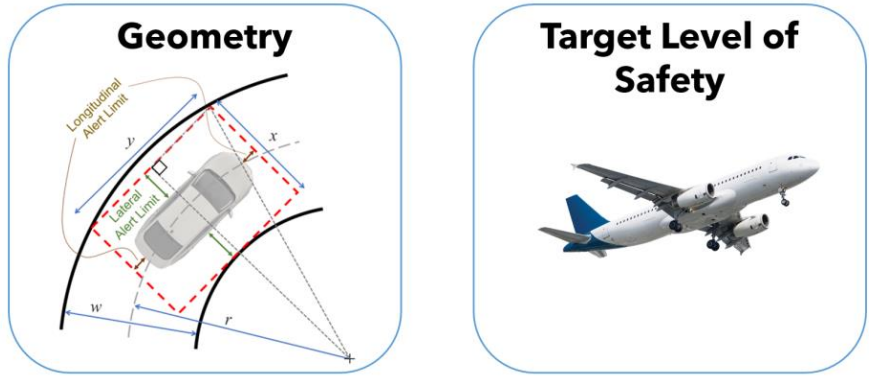

*Figure 2: Localization and mapping requirements are governed by (1) geometry to define the bounding box required for different operations (levels of situational awareness) and (2) the target level of safety to define the requisite accuracy and integrity.*

**LOCALIZATION & MAPPING GEOMETRY**

Situational awareness requirements in highly automated and fully autonomous vehicles are generally grouped into the following categories: (1) Road determination; (2) Lane determination; and (3) In-lane positioning. Each has their function which ranges from geofencing to path planning to lane keeping as discussed in (Reid, Pervez, et al., 2019). Their relative difference is shown pictorially in Figure 3 which also indicates the approximate position error usually associated with these applications. In this section, we derive a more formalized geometric bounding box required for each. The horizontal geometry is dictated by the road width and curvature along with vehicle size standards (Reid, Houts, et al., 2019). Vertical requirements are governed by the interspacing between roads amongst multilevel interchanges. We will focus on road and vehicle standards in the United States though the methods are general and can be applied in other regions.

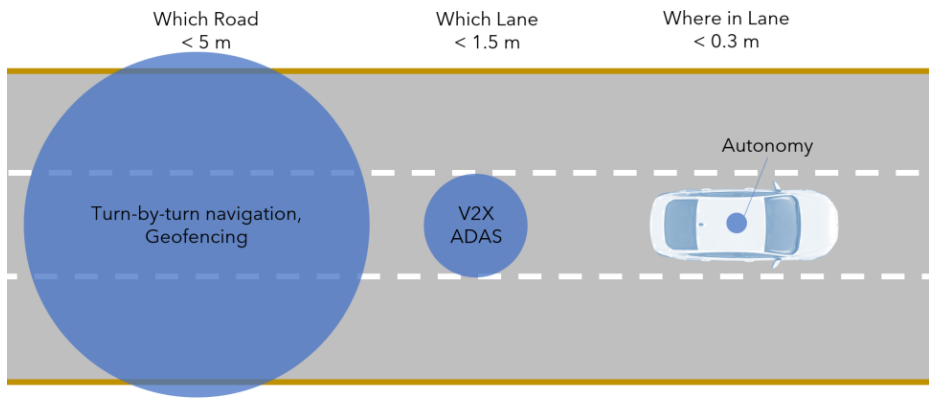

*Figure 3: The three general levels of location situational awareness required in automotive applications and the approximate positioning levels usually associated with them.*

Vehicle dimensions standards in the United States are shown in Table 1. We will focus on passenger vehicles though it is noteworthy that passenger vehicle dimension limits include even six wheeled 'dualie' pickup trucks and hence represents a broad range of vehicles. Table 2 shows road geometry. Road curvature is a function of design speed and is based on limiting values of side friction factor and superelevation (American Association of State Highway and Transportation Officials, 2001). Superelevation is the rotation of the pavement on the approach to and throughout a horizontal curve and is intended to help the driver by countering the lateral acceleration produced by tracking the curve. The other important factor is road width, which typically ranges from 3.6 meters on standard freeways to 2.7 meters on limited residential streets. Along with vehicle dimensions, road width and curvature are the elements that define the bounding box for the situational awareness levels of interest. The limiting cases for each road type have been assembled in Table 2 for passenger vehicles.

*Table 1: Vehicle dimension standards in the United States (American Association of State Highway and Transportation Officials, 2001; U.S. Department of Transportation Federal Highway Administration, 2017).*

| Vehicle Type | Width [m] | Length [m] | Height [m] |
|---|---|---|---|
| Passenger | 2.1 | 5.8 | 1.3 |
| Single Unit Truck | 2.4 | 9.2 | 3.4 – 4.1 |
| City Bus | 2.6 | 12.2 | 3.2 |
| Semitrailer | 2.4 – 2.6 | 13.9 – 22.4 | 4.1 |

*Table 2: Limiting road design elements for passenger vehicles in the United States (American Association of State Highway and Transportation Officials, 2001; United States Department of Transportation & Federal Highway Administration, 2000; Washington State Department of Transportation, 2017).*

| Road Type | Design Speed [km/h] | Lane Width [m] | Minimum Radius [m] |
|---|---|---|---|
| Freeway | 80 – 130 | 3.6 | 195[**] |
| Interchanges | 30 – 110 | 3.6 – 5.4 | 150 – 15 |
| Arterial | 50 – 100 | 3.3 – 3.6 | 70[**] |
| Collector | 50 | 3.0 – 3.6 | 70[**] |
| Local | 20 – 50 | 2.7[*] – 3.6 | 10[**] |
| Hairpin Turn / Cul-de-Sac | < 20 | 6.0 | 7 |
| Single Lane Roundabout | < 20 | 4.3 | 11 |

[*]The lower bound of 2.7 m is the exception, not the rule, and is typically reserved for residential streets with low traffic volumes.
[**]Based on design speeds and limiting values of rate of roadway superelevation and coefficient of friction.

Following the methodology of Reid, Houts, et al., (2019), we select a bounding box to derive position alert limits. This is a function of the road geometry, vehicle dimensions, and desired level of situational awareness for the intended function. Other shapes, such as elliptical geometry has been shown to somewhat relax alert limits (Feng et al., 2018; Kigotho & Rife, 2021). In this analysis, we select a bounding box due in part to simplicity but also to its conservative nature compared to other shapes. The bounding box geometry for (1) lane keeping, (2) lane determination, and (3) road determination is shown in Figure 4. This picture allows us to conceptually extend the geometric relationships developed by Reid, Houts, et al., (2019) towards these levels of situational awareness. The result is summarized in Table 3. This represents the allowable bounds that result from the combined position, attitude, and map errors of the collective system. This inherits many of the assumptions of Reid, Houts, et al., (2019). Beginning with local roads, relatively sharp turns (small radius of curvature) necessitate similar longitudinal and lateral errors due to the coupling that results from the curvature of the road. The result is 0.33 m in both lateral and longitudinal for lane keeping, 1.38 m lateral and 3.23 m longitudinal for lane determination, and 2.89 m lateral and 3.23 m longitudinal for road determination, considered here as 2 lanes. For freeway road geometry, we inherit the assumption from Reid, Houts, et al., (2019) that useful situational awareness comes at the level of half a vehicle length. This is not only to use road curvature information from a map for steering or for finding on/off ramps, but also for V2X applications where vehicles share information towards a broader group situational awareness for improved safety. Building on this assumption, the result is 0.72 m lateral and 1.50 m longitudinal for lane keeping, 1.77 m lateral and 4.40 m for lane determination, and 3.57 m lateral and 4.40 m longitudinal for road determination, again considered as two lanes. In all cases, we assume the same level of vertical bounds at 1.47 m. This inherits the argument from Reid, Houts, et al. (2019) and is based on the vertical separation between road decks. Standards in the U.S. dictate this clearance to be 4.4 m, however, knowledge to half this separation is insufficient for road deck resolution amongst multi-level interchanges such as the 'High-Five' interchange in Dallas, TX, famous for its five decks. Following Reid, Houts, et al. (2019), we will conservatively choose 1/3 of this separation or 1.47 m.

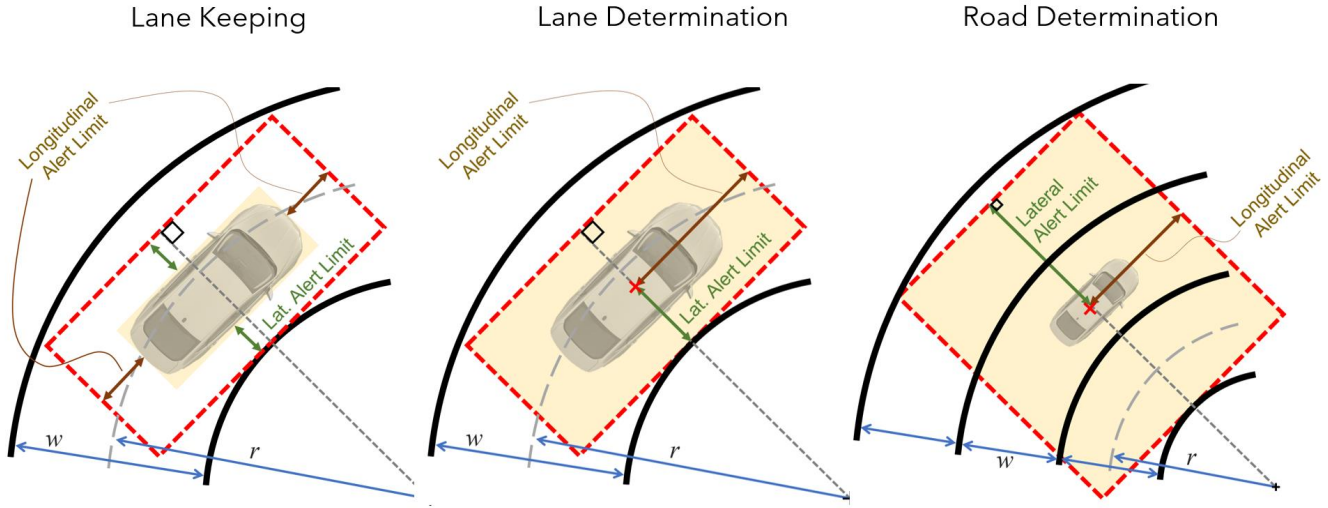

*Figure 4: Comparison of bounding box geometry for (1) Lane keeping; (2) Lane determination; and (3) Road determination. Road determination in this picture is shown for a 3-lane freeway as an example.*

*Table 3: Horizontal (lateral / longitudinal) and vertical alert limits for U.S. local roads and freeways based on a bounding box geometry.*

| Situational Awareness Level | Local Roads | | | Freeways & Interchanges | | |
|---|---|---|---|---|---|---|
| | Lateral [m] | Longitudinal [m] | Vertical [m] | Lateral [m] | Longitudinal [m] | Vertical [m] |
| Road Determination* | 2.89 | 3.23 | 1.47 | 3.57 | 4.40 | 1.47 |
| Lane Determination | 1.38 | 3.23 | 1.47 | 1.77 | 4.40 | 1.47 |
| Lane Keeping | 0.33 | 0.33 | 1.47 | 0.72 | 1.50 | 1.47 |

*Road determination is considered as two lanes in this analysis.

**INTEGRITY FOR ADAS**

In this section, we develop an integrity risk model for ADAS and its subsystems which includes the human-driver-in-the-loop. This expands on previous work which focused on localization and integrity for Level 4+ full self-driving which did not include human factors (Reid, Houts, et al., 2019). To develop the ADAS model, we draw analogy to civil aviation where the pilot is considered in-the-loop for certain operations akin to the Level 2+ use case. We also present some lessons learned from aviation that may be considered towards safety certification of automated driving systems. We also make recommendations on the initial Target Level of Safety (TLS) for ADAS based on the human element of the system.

**Analogy to Civil Aviation**

It has been argued that the long-term safety goal of Level 4+ full self-driving systems be the same as other forms of mass transportation, where the gold standard today is civil aviation (Reid, Houts, et al., 2019). This represents an on-road equivalency on the order of one fatality per ten billion miles travelled where today this number is closer to one in one hundred million miles. In civil aviation, automated precision approach and landing systems have been a major component towards achieving safety goals. First developed during World War II, these systems were introduced into civil aviation in the 1960s and gained widespread adoption in the 1980s, enabling instrument-based aircraft control when visibility is poor (Charnley, 2011). As a result, the aviation industry has pioneered the development of navigation systems for safety critical operations in automated human transportation. Alongside this, civil aviation has developed processes for organizing the international community for developing standards and certification methodology.

In developing GPS-based precision approach techniques, the system level standards were established by the International Civil Aviation Organization (ICAO), a specialized agency within the United Nations. The ICAO defines the degree of position certainty and integrity required in specific stages of flight to achieve the desired level of safety. These standards necessitated development by an international organization as aircraft must interoperate across political borders. In the United States, the technology solution was developed in-part as a government service, where the U.S. Federal Aviation Administration (FAA) deployed the Wide Area Augmentation System (WAAS), a GPS Space-Based Augmentation Service (SBAS) whose mission is to overlay the information needed to achieve safety in using GPS for aircraft navigation (Enge et al., 1996). The other side of the technology solution was end user equipment – the GPS receiver on the aircraft. To achieve integrity and safety certification, implementation standards were established by the Radio Technical Commission for Aeronautics (RTCA) which brought together the aviation industry, WAAS service provider, and GPS receiver manufactures to form the so-called Minimum Operational Performance Standard (MOPS) which is an implementation cookbook on how to use the WAAS service to achieve integrity, allowing multiple manufacturers to build the user equipment (Radio Technical Commission for Aeronautics, 2020). This defines clear expectations and responsibilities of each player so that when location services are utilized by user equipment, safety certification can be achieved. These elements are summarized in Figure 5.

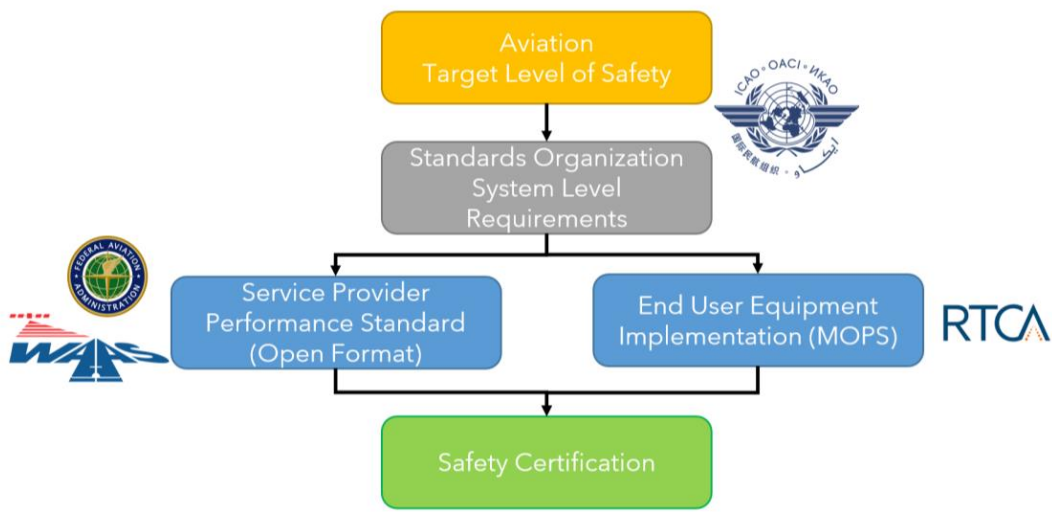

*Figure 5: Process of GPS-based navigation system safety certification in civil aviation.*

In developing an integrity risk model for automated driving, we consider two flight operations which offer analogous elements as summarized in Table 4. The first is a Category III (Cat III) landing which is fully automated, akin to SAE Level 4+ fully autonomous vehicles. Sensor information provides the situational awareness needed for full automated control of the aircraft, completing the final landing maneuver. The speeds involved during this operation result in timescales that are too short for the human pilot to intervene and hence the pilot does not contribute to the safety case. This is highlighted in the Cat III integrity risk tree in Figure 6 which shows the assumed pilot's ability to detect failures for a go-around and second landing attempt as a 1 in 1 failure, indicating no ability to intervene. The second operation considered is semi-autonomous which supports a Cat I precision runway approach, akin to SAE Level 2+ partial autonomy. In this scenario, the automated system brings the aircraft down to a decision height of 200 ft (60 m) where the pilot makes the final decision to commit to landing. It is further assumed that the pilot has sufficient time to assess the situation and resume manual control if there is a recognized system fault before the final decision height is reached. This is not unlike a disengagement made by a human driver in a Level 2+ partially automated vehicle. In Cat I, the pilot is indicated as having a 25% chance of recognizing a system fault and hence contributes to safety case as shown in the risk tree in Figure 7. This approach to the inclusion of the human-in-the-loop in Cat I will form the basis of our ADAS integrity risk model.

Table 4: Comparison of automated landing categories in civil aviation (Roturier et al., 2001; Speidel et al., 2013). CAT I assumes pilot oversight on precision approach while CAT III is fully autonomous touch down in zero visibility conditions.

| Aircraft Automated Landing Type | Decision Height | Visibility Req. | Pilot Role | Accuracy (95% or 1.96σ) | | Alert Limit (5 – 6 σ) | | Probability of Failure (Integrity) | Analogous SAE Self-Driving Category |
|---|---|---|---|---|---|---|---|---|---|
| | | | | Horizontal | Vertical | Horizontal | Vertical | | |
| CAT I | 60 m (200') | > 550 m (>1800') | Monitor for faults, take control if necessary | 16 m | 4 m | 40 m | 10 m | $10^{-7}$ / approach | Level 2 |
| CAT III | No Limit | None | None, fully autonomous | 6.9 m | 2.9 m | 17 m | 5.3 m | $10^{-9}$ / approach | Level 4 |

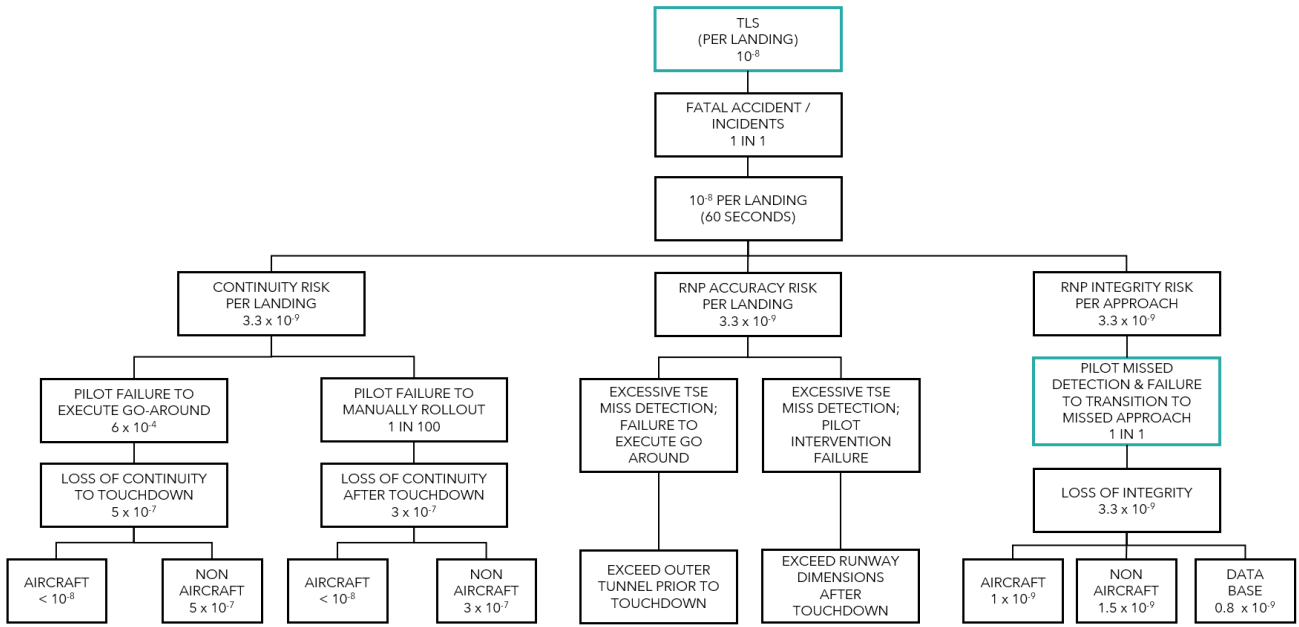

Figure 6: Category III automated aircraft landing system risk allocation. This is full automation, where the pilot is assumed to have no input during the final touchdown. This highlights the Target Level of Safety (TLS) and the pilot's contribution to integrity (highlighted in teal), showing an assumed 1 in 1 chance of failed detection of system faults. Based on (Kelly & Davis, 1994).

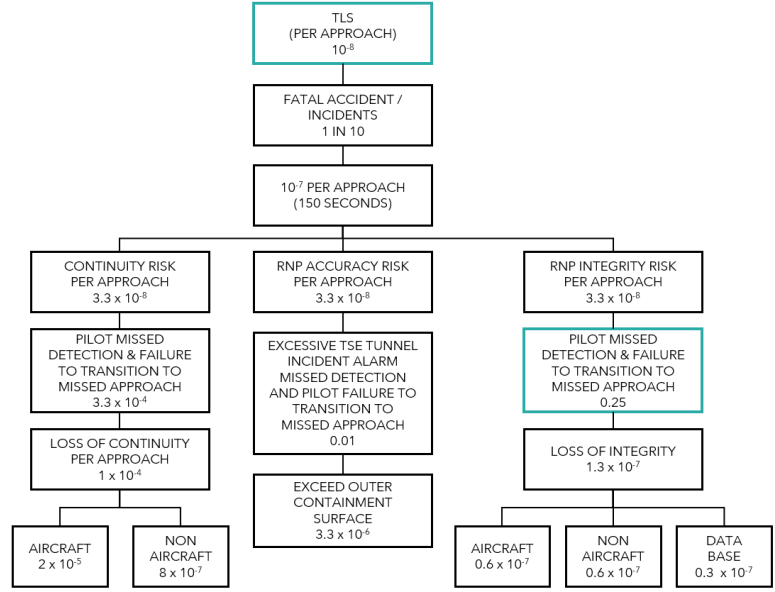

*Figure 7: Category I automated aircraft precision approach system risk allocation. This partially automated system brings the aircraft to a decision height of 200 ft (60 m) above the runway for final decision of the pilot. This highlights the Target Level of Safety (TLS) and the pilot's contribution to integrity (highlighted in teal), showing an assumed 1 in 4 chance of failed detection of system faults. Based on (Kelly & Davis, 1994).*

**ADAS & Autonomous Vehicles**

Integrity risk allocation for Level 4+ autonomous vehicle systems was developed by Reid, Houts, et al. (2019) where Figure 8 shows a simplified block diagram of the integrity risk model developed. At the core of Figure 8 is the Target Level of Safety (TLS), a concept borrowed from civil aviation. This shows a TLS equivalent to that achieved in commercial aviation, the target suggested as the long-term safety goal for automated road vehicles (Reid, Houts, et al., 2019). In this framework, the first step to determining allowable integrity risk for subsystems is to work backwards from the TLS and understand accident survivability. The fatal accident to incident ratio $P_{F:I}$ accounts for the fact that not all accident causing system failures directly result in a fatality. As shown in Figure 7, in a Cat I precision aircraft approach $P_{F:I}$ is taken as 1/10 or one fatal accident per ten total accidents or incidents. However, it has been shown that automotive accidents are generally more survivable due to the lower energies involved, i.e. lower velocities and impossibility of falling to the ground, and a more appropriate rate is 1/100 fatal accidents per total accidents (Reid, Houts, et al., 2019). In the analysis of Level 4+ self-driving systems, the resulting system level integrity risk (or probability of system failure) is assumed to be shared equally between vehicle systems $P_{veh}$ and the virtual driver systems $P_{vds}$. Data suggests vehicle systems to very nearly be at the level of $10^{-8}$ failures / mile today and hence future self-driving features should be held to the same standard towards achieving overall safety goals (Reid, Houts, et al., 2019). The equation relating these elements is as follows:

$$TLS = P_{F:I}(P_{veh} + P_{vds}) \qquad (1)$$

The detailed systems analysis by Reid, Houts, et al. (2019) suggests that the localization subsystem will only be allocated a small fraction of the overall virtual driver system budget, as it is assumed that errors cascade downstream from localization to control. The result is that localization is allotted approximately ten percent of the $P_{vds}$ budget, resulting in $P_{loc} = 0.1 \times P_{vds} = 10^{-9}$ failures / mile. This is among the most stringent requirements of virtual driver subsystems, representing one failure in a billion miles, which is 250x the entire U.S. road network.

Here, we adapt the SAE Level 4+ model from Reid, Houts, et al. (2019) to SAE Level 2+ ADAS which includes human driver oversight. For comparison, the first step is to develop a model for today's human driver and vehicle systems. Figure 9 shows a data driven model for vehicles and drivers in the United States based on road safety statistics from the U.S. National Highway Traffic Safety Administration (NHTSA) (National Highway Traffic Safety Administration & US Department of Transportation, 2020; Singh, 2015). In this instance, the level of safety is an observed quantity, not a TLS. This reflects the state of road safety where the achieved level in 2019 was 1.11 fatalities per 100 million vehicle mile traveled (National Highway Traffic Safety Administration & US Department of Transportation, 2020). As the number of vehicle occupants is often more than one, this corresponds to 1.02 fatal crashes per 100 million vehicle miles traveled.

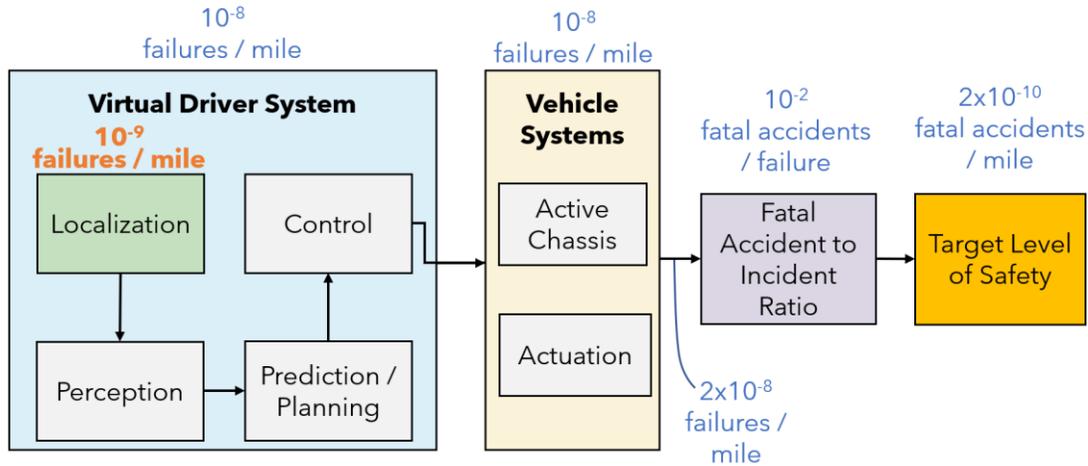

*Figure 8: Simplified integrity risk allocation for a Level 4+ self-driving system. This shows the integrity risk allocation of the localization subsystem required to meet the desired Target Level of Safety (TLS) akin to that achieved in civil aviation. (Reid, Houts, et al., 2019).*

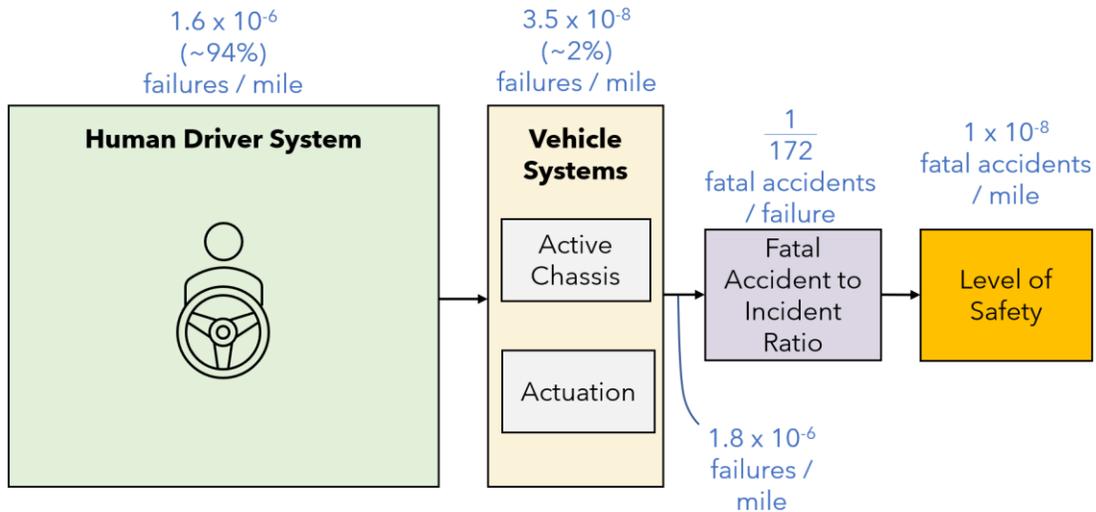

*Figure 9: Data-driven model of road vehicle systems today. This includes a model of the performance of both vehicle systems and the human driver in terms of probability of failure per mile.*

The fatal accident to incident ratio $P_{F:I}$ will also be taken directly from the data, which was estimated to be 1/172 fatal car crashes per total crashes on average between 1990 and 2015 (Reid, Houts, et al., 2019). Crash causation statistics allow us to estimate the relative contribution of the human driver, vehicle systems, and other factors and are estimated by NHTSA as follows: 94%(±2.2%) from human driver errors; 2%(±0.7%) from electrical and mechanical component failures; 2%(±1.3%) from environmental factors contributing to slick (low μ) roads such as water, ice, snow, etc.; and 2%(±1.4%) from unknown reasons (National Highway Traffic Safety Administration & US Department of Transportation, 2020). Working backwards from the observed level of safety achieved, we can estimate the integrity risk of the human driver system $P_{hds}$ and vehicle systems $P_{veh}$ as follows:

$$P_{veh} = 0.02 \frac{TLS}{P_{F:I}} = 0.02 \frac{1.02 \times 10^{-6} \text{ fatal accidents/mile}}{1/172 \text{ fatal accidents/failure}} \approx 3.5 \times 10^{-8} \frac{\text{failures}}{\text{mile}} \quad (2)$$

$$P_{hum} = 0.94 \frac{TLS}{P_{F:I}} = 0.94 \frac{1.02 \times 10^{-6} \text{ fatal accidents/mile}}{1/172 \text{ fatal accidents/failure}} \approx 1.6 \times 10^{-6} \frac{\text{failures}}{\text{mile}} \quad (3)$$

This forms the performance baseline for a human driver system today as summarized in Figure 9. With the long-term Level 4+ full self-driving goals stated above, this targeted hundredfold improvement in road safety forms the gap to be ultimately closed with ADAS, active safety, and autonomous features.

For ADAS, we must combine elements of the virtual driver system with a human driver in the loop whose role is to provide oversight. To develop this model, we will again borrow from civil aviation, particularly a Cat I automated landing system and the roll of the human pilot. In Cat I, the automated system brings the aircraft down to a decision height of 200 ft (60 m) at which point it is up to the judgement of the pilot to take over control and restart the approach or to follow through and land the aircraft (Speidel et al., 2013). The role of the pilot is to constantly monitor the automated system for faults and take over, if necessary, akin to a disengagement in SAE Level 2+ automated road vehicles. As shown in Figure 7, the pilot is assumed to have a misdetection rate of 25% and is included in the integrity risk allocation calculations as part of the system design. For a human driver, studies indicate that takeover failure rates are approximately of 37.2% (Yu et al., 2021). In our ADAS integrity risk model, we will conservatively assume a driver oversight misdetection (dom) rate of $P_{dom} = 40\%$.

Figure 10 shows the resulting ADAS integrity risk model including the resulting TLS and human driver in the loop. The elements are related as follows to include the influence of the driver:

$$TLS = P_{F:I}(P_{veh} + P_{dom}P_{ADAS}) \quad (4)$$

where $P_{ADAS}$ is the probability of failure of the ADAS semi-autonomous driving system in its Operational Design Domain (ODD). In Figure 10 we show the TLS assuming the ADAS block is designed to the same level of performance as a human driver, where $P_{ADAS} = P_{hds} \approx 1.6 \times 10^{-6}$ failures / mile. The result is a TLS = $4 \times 10^{-9}$ fatal accidents / mile, or a 2.5x improvement in road safety compared to today. It is interesting to note that Tesla's self-reported accident statistics indicate approximately a 2.5x improvement in road safety when the Tesla Level 2+ Autopilot is engaged compared to when Tesla vehicles are human driven without any active safety features (Tesla, 2021). The Tesla accident statistics are summarized in Table 5.

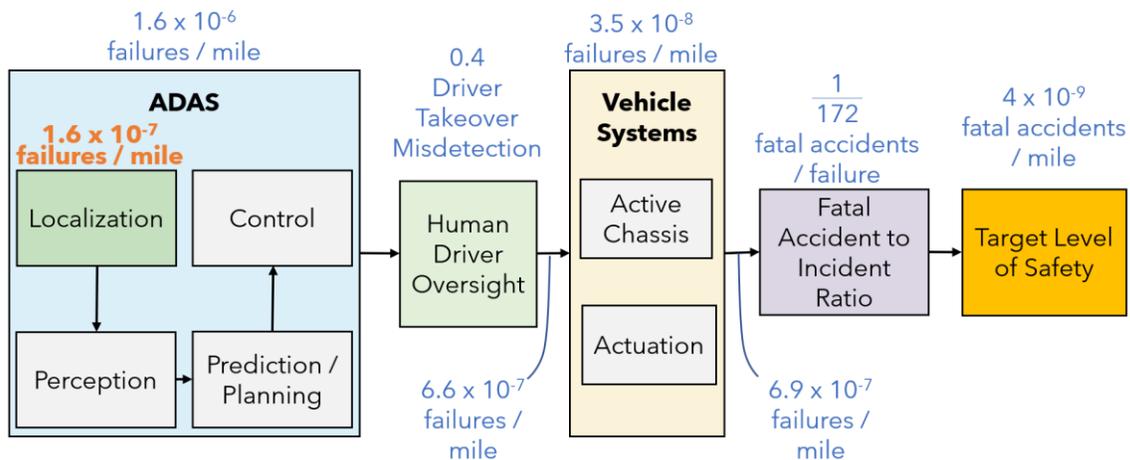

*Figure 10: Integrity risk allocation model for ADAS. This includes the human driver in the loop serving an in an oversight role. This shows an example of an ADAS system designed to the level of performance of a human driver in its Operational Design Domain (ODD).*

Table 5: Summary of Tesla self-reported accident data (Tesla, 2021). This shows approximately a 2-3x increase in road safety for vehicles driven with Tesla's Autopilot compared to no active safety features.

| Data Period | Millions of Miles Between Accidents | | | | Accidents / Mile | | | | Accident Ratio of Autopilot to No Safety Features |
|---|---|---|---|---|---|---|---|---|---|
| | NHTSA | Tesla No Safety Features | Tesla Active Safety Features | Tesla Autopilot | NHTSA | Tesla No Safety Features | Tesla Active Safety Features | Tesla Autopilot | |
| Q3 2018 | 0.492 | 2.02 | 1.92 | 3.34 | 2.0E-06 | 5.0E-07 | 5.2E-07 | 3.0E-07 | 1.65 |
| Q4 2018 | 0.436 | 1.25 | 1.58 | 2.91 | 2.3E-06 | 8.0E-07 | 6.3E-07 | 3.4E-07 | 2.33 |
| Q1 2019 | 0.436 | 1.26 | 1.76 | 2.87 | 2.3E-06 | 7.9E-07 | 5.7E-07 | 3.5E-07 | 2.28 |
| Q2 2019 | 0.498 | 1.41 | 2.19 | 3.27 | 2.0E-06 | 7.1E-07 | 4.6E-07 | 3.1E-07 | 2.32 |
| Q3 2019 | 0.498 | 1.82 | 2.7 | 4.34 | 2.0E-06 | 5.5E-07 | 3.7E-07 | 2.3E-07 | 2.38 |
| Q4 2019 | 0.479 | 1.64 | 2.1 | 3.07 | 2.1E-06 | 6.1E-07 | 4.8E-07 | 3.3E-07 | 1.87 |
| Q1 2020 | 0.479 | 1.42 | 1.99 | 4.68 | 2.1E-06 | 7.0E-07 | 5.0E-07 | 2.1E-07 | 3.30 |
| Q2 2020 | 0.479 | 1.56 | 2.27 | 4.53 | 2.1E-06 | 6.4E-07 | 4.4E-07 | 2.2E-07 | 2.90 |
| Q3 2020 | 0.479 | 1.79 | 2.42 | 4.59 | 2.1E-06 | 5.6E-07 | 4.1E-07 | 2.2E-07 | 2.56 |
| Q4 2020 | 0.484 | 1.27 | 2.05 | 3.45 | 2.1E-06 | 7.9E-07 | 4.9E-07 | 2.9E-07 | 2.72 |

Following the system breakdown and integrity risk allotment described by Reid, Houts, et al. (2019), only 10% of the $P_{ADAS}$ budget will be allocated to the localization subsystem. This results in $P_{loc} = 0.1 \times P_{ADAS} = 1.6 \times 10^{-7}$ failures / mile. This is the probability of the localization subsystem outputting Hazardous Misleading Information (HMI) which is in turn consumed by downstream systems including perception, planning, and control. To get this in terms of failures per hour of operation (a more common unit), we follow the vehicle speed range described by Reid, Houts, et al. (2019) where lower bounds are taken as the minimum speed at which airbags will deploy, corresponding to 10 mph (16 km/h) (Wood et al., 2014). The result is as follows:

$$P_{loc} = 1.6 \times 10^{-7} \frac{\text{failures}}{\text{mile}} \times 10 \frac{\text{mile}}{\text{hour}} \approx 10^{-6} \frac{\text{failures}}{\text{hour}} \quad (5)$$

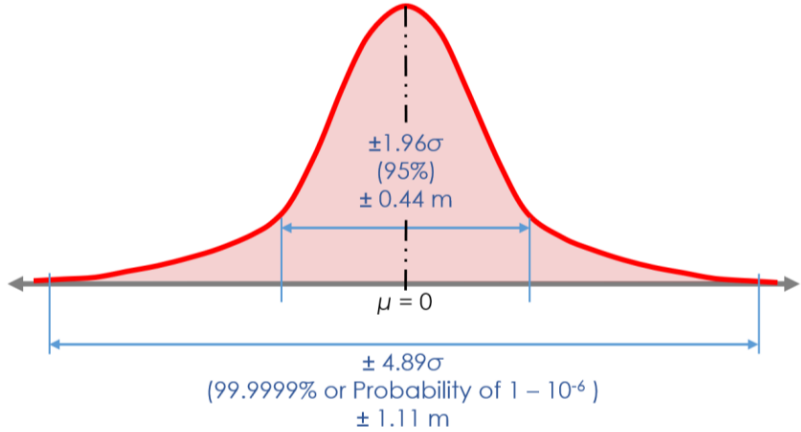

Figure 11: The desired error distribution for lateral positioning on freeways for passenger vehicle dimension limits, assuming a Gaussian distribution. This shows the 95% accuracy at 0.44 meters and error bounds at 1.11 m at 99.9999% which is a probability of (1 – $10^{-6}$). This assumes equal allocation of between the mapping and localization systems.

To understand the implied error distribution for the desired level of integrity, Figure 11 shows the resulting Gaussian that results from 1 – $10^{-6}$ or 99.9999% of lateral positions being within the acceptable bounds of 1.11 meters for highway operation. For a Gaussian distribution with mean $\mu$ (assumed zero) and standard deviation $\sigma$, we use the error function to determine the expected fraction of occurrences outside of the range $\mu \pm N\sigma$. Using the target of $10^{-6}$ failures per hour, we can determine $N$ as follows:

$$\text{erf}\left(\frac{N}{\sqrt{2}}\right) = 1 - 10^{-6} \quad (6)$$

$$N_{(1-10^{-6})} = \sqrt{2}\ \text{erf}^{-1}(1 - 10^{-6}) = 4.89 \tag{7}$$

The corresponding 95% error bounds are computed similarly:

$$N_{0.95} = \sqrt{2}\ \text{erf}^{-1}(0.95) = 1.96 \tag{8}$$

where the 95% position accuracy expected for such a Gaussian distribution is $1.96/4.89 \times 1.11\ \text{m} = 0.44\ \text{m}$.

To put probability of failure per hour of operation in context, we will compare it to safety standards across different industries. Table 6 shows an approximate cross-domain mapping of aviation, rail, general programmable electronics, and automotive safety integrity levels alongside their corresponding probability of failure per hour targets. In rail, aviation, and programmable electronics, the strictest levels are those corresponding to failures which could cause the loss of multiple human lives and correspond to an integrity level of $10^{-9}$ failures / hour. In rail and electronics this is Safety Integrity Level (SIL) 4. In aviation, this is Design Assurance Level (DAL) A. The automotive industry's strictest requirement, Automotive Safety Integrity Level (ASIL) D is closer to SIL-3 and DAL-B in practice or in the range of $10^{-8}$ to $10^{-7}$ failures / hour. Though not adequate to fully capture autonomous driving needs, when referring to probability of failure per hour of operation, it is still a useful exercise to compare to those of ISO 26262. This places localization for ADAS in the ASIL B range by this metric. ASIL B has emerged as generally the target of many ADAS systems, but this provides a different perspective based on first principles as to why this is a sensible target.

*Table 6: Approximate cross-domain mapping of safety integrity levels (Baufreton et al., 2010; Blanquart et al., 2012; Kafka, 2012; Machrouh et al., 2012; Verhulst, 2013; Verhulst & Sputh, 2013).*

| Probability of Failure per Hour | General Programmable Electronics IEC-61508 | Automotive ISO 26262 | Aviation DO-178/254 | Railway CENELEC 50126 128/129 |
|---|---|---|---|---|
| - | (SIL-0) | QM | DAL-E | (SIL-0) |
| $10^{-6} - 10^{-5}$ | SIL-1 | ASIL-A | DAL-D | SIL-1 |
| $10^{-7} - 10^{-6}$ | SIL-2 | ASIL-B/C | DAL-C | SIL-2 |
| $10^{-8} - 10^{-7}$ | SIL-3 | ASIL-D | DAL-B | SIL-3 |
| $10^{-9} - 10^{-8}$ | SIL-4 | - | DAL-A | SIL-4 |

**MAPPING & LOCALIZATION BUDGET**

The bounding box geometry given in Table 3 represents the combined error budget from position, attitude (orientation), as well as from the map. This section will present a framework for separating the map and localization components from the combined error budget.

Following the method of Joubert et al. (2020), the relationship between the map and localization error budget can be written as:

$$\sigma_{total}^2 = \sigma_{loc}^2 + \sigma_{map}^2 \tag{9}$$

where $\sigma_{loc}$ is the standard deviation of the localization system, $\sigma_{map}$ is the standard deviation of the map, and $\sigma_{total}$ is the total budget between them. If we allow equal allocation of error between the mapping and localization systems, $\sigma_{loc} = \sigma_{map} = \sigma_{alloc}$, we obtain:

$$\sigma_{total}^2 = 2\sigma_{alloc}^2 \tag{10}$$

solving for this allocation gives:

$$\sigma_{alloc} = \frac{\sigma_{total}}{\sqrt{2}} \tag{11}$$

Hence, if the desired map + localization error distribution is known, dividing any $N\sigma_{total}$ by $\sqrt{2}$ will result in the allocation for both components, i.e. $N\sigma_{loc} = N\sigma_{map} = N\sigma_{total}/\sqrt{2}$.

**LOCALIZATION & MAPPING REQUIREMENTS SUMMARY**

In this section, we summarize recommendations on requirements for position, attitude (orientation), and map georeferencing for different levels of situational awareness to support Level 2+ highly automated vehicles. Road determination, lane determination, and lane keeping each unlock new features. For example, road determination can geofence features with partial autonomy to limited access divided highways. Lane determination has been suggested to be a significant step towards unlocking the next generation of features for ADAS and enabling true hands off / eyes off driving in certain ODDs (Joubert et al., 2020). Lane keeping levels of localization represents the level of situational awareness required for full Level 4 self-driving (Reid, Houts, et al., 2019). Combining the bounding box geometry and target level of safety results in the joint localization and mapping error budget summarized in Table 7 and Table 8 for U.S. freeways and local roads, respectively. As described by Reid, Houts, et al. (2019), there is not one unique set of requirements but rather a family that satisfy the bounding box derived in Table 3, allocating between lateral, longitudinal, and vertical positioning as well as attitude (orientation). Table 7 and Table 8 represent a design point that satisfies the bounds given in Table 3, though others are possible.

*Table 7: Combined localization + map error budget for U.S. freeway operation with interchanges. This assumes minimum lane widths of 3.6 meters and allowable speeds up to 137 km/h (85 mph).*

| Situational Awareness Level | Accuracy (95%) | | | | Alert Limit | | | | Prob. of Failure (Integrity) [per hour] |
|---|---|---|---|---|---|---|---|---|---|
| | Lateral [m] | Long. [m] | Vertical [m] | Attitude [deg] | Lateral [m] | Long. [m] | Vertical [m] | Attitude [deg] | |
| Road Determination (2 Lanes) | 1.28 | 1.63 | 0.56 | 1.60 | 3.19 | 4.08 | 1.40 | 4.00 | 1.0E-06 |
| Lane Determination | 0.63 | 1.72 | 0.56 | 0.80 | 1.57 | 4.30 | 1.40 | 2.00 | 1.0E-06 |
| Lane Keeping | 0.23 | 0.57 | 0.56 | 0.60 | 0.58 | 1.43 | 1.40 | 1.50 | 1.0E-06 |

*Table 8: Combined localization + map error budget for U.S. local roads. This assumes minimum lane widths of 3.0 meters with a minimum curvature of 20 meters or 3.3 meters with a minimum curvature of 10 meters.*

| Situational Awareness Level | Accuracy (95%) | | | | Alert Limit | | | | Prob. of Failure (Integrity) [per hour] |
|---|---|---|---|---|---|---|---|---|---|
| | Lateral [m] | Long. [m] | Vertical [m] | Attitude [deg] | Lateral [m] | Long. [m] | Vertical [m] | Attitude [deg] | |
| Road Determination (2 Lanes) | 1.09 | 1.24 | 0.56 | 0.80 | 2.73 | 3.10 | 1.40 | 2.00 | 1.0E-06 |
| Lane Determination | 0.50 | 1.26 | 0.56 | 0.60 | 1.26 | 3.15 | 1.40 | 1.50 | 1.0E-06 |
| Lane Keeping | 0.12 | 0.12 | 0.56 | 0.20 | 0.29 | 0.29 | 1.40 | 0.50 | 1.0E-06 |

Assuming the map is allocated the same level of error budget as real-time localization as described in the previous section, we arrive at the final set of SAE Level 2+ localization and map georeferencing requirements summarized in Table 9 and Table 10 for U.S. freeways and local roads, respectively. This may seem unfair as mapping can be computed offline, but such a breakdown enables fleet vehicles to act as mapping vehicles running real-time localization schemes at similar levels of performance. These summaries break down localization and map geometry requirements by 95% (1.96 σ) accuracy and alert limits or the maximum allowable bounds at an integrity risk of $10^{-6}$ (4.89 σ). Combined, these describe a desired Gaussian distribution of errors in lateral, longitudinal, and vertical positioning along with attitude (orientation). Notice that 95% accuracy in lateral is less than 1 meter even for 2-lane lane freeway road determination while local road lane keeping is at the level of 8 centimeters, representing nearly an order of magnitude difference. Herein represents the challenge in highly automated driving systems where different features can require vastly different levels of localization to support their intended function and hence the challenge going forward.

*Table 9: Localization and map georeferencing requirements for U.S. freeway operation with interchanges. This assumes minimum lane widths of 3.6 meters and allowable speeds up to 137 km/h (85 mph). This assumes equal allocation has been allotted between the localization and map components.*

| Situational Awareness Level | Accuracy (95%) | | | | Alert Limit | | | | Prob. of Failure (Integrity) |
|---|---|---|---|---|---|---|---|---|---|
| | Lateral [m] | Long. [m] | Vertical [m] | Attitude [deg] | Lateral [m] | Long. [m] | Vertical [m] | Attitude [deg] | [per hour] |
| Road Determination (2 Lanes) | 0.90 | 1.16 | 0.40 | 1.13 | 2.26 | 2.88 | 0.99 | 2.83 | 1.0E-06 |
| Lane Determination | 0.44 | 1.22 | 0.40 | 0.57 | 1.11 | 3.04 | 0.99 | 1.41 | 1.0E-06 |
| Lane Keeping | 0.16 | 0.41 | 0.40 | 0.43 | 0.41 | 1.01 | 0.99 | 1.06 | 1.0E-06 |

*Table 10: Localization and map georeferencing requirements for U.S. local roads. This assumes minimum lane widths of 3.0 meters with a minimum curvature of 20 meters or 3.3 meters with a minimum curvature of 10 meters. This assumes equal allocation has been allotted between the localization and map components.*

| Situational Awareness Level | Accuracy (95%) | | | | Alert Limit | | | | Prob. of Failure (Integrity) |
|---|---|---|---|---|---|---|---|---|---|
| | Lateral [m] | Long. [m] | Vertical [m] | Attitude [deg] | Lateral [m] | Long. [m] | Vertical [m] | Attitude [deg] | [per hour] |
| Road Determination (2 Lanes) | 0.77 | 0.88 | 0.40 | 0.57 | 1.93 | 2.19 | 0.99 | 1.41 | 1.0E-06 |
| Lane Determination | 0.36 | 0.89 | 0.40 | 0.43 | 0.89 | 2.23 | 0.99 | 1.06 | 1.0E-06 |
| Lane Keeping | 0.08 | 0.08 | 0.40 | 0.14 | 0.21 | 0.21 | 0.99 | 0.35 | 1.0E-06 |

**SYSTEM DECOMPOSITION**

Achieving the requirements in Table 9 and Table 10 will likely require a combination of technologies to achieve the desired performance. Figure 12 shows a common architecture emerging in the autonomous vehicle industry based on redundant systems working in tandem (Joubert et al., 2020). This comprises a relative localization system based on cameras, radar, or lidar alongside an absolute localization system comprised of Global Navigation Satellite Systems (GNSS). The relative approach relies on matching local sensor data to an a-priori map and hence navigation is relative to referenced objects. This method tends to have challenges in sparse open or repetitive environments as feature sets are not sufficiently unique to compute a position, such as open or rural highways. By contrast, the absolute localization approach leverages satellite navigation, which can face challenges in areas with sky obstructions such as dense urban canyons. In both cases, inertial systems provide position information during outages and typically include additional velocity inputs to limit the inertial drift. At face value, these are complementary. Relative sensors use GNSS obstructions as navigation landmarks and GNSS works best in open sparse environments.

In addition to having independent failure modes, in order for a combination of localization systems to satisfy overall requirements, they must each meet performance thresholds. One approach is to leverage system decomposition in the ISO 26262 standard, which is one of many being adapted for automated driving. For example, an ASIL D system can be decomposed into two independent ASIL B systems and likewise, an ASIL B system can be decomposed into two ASIL A systems (Kafka, 2012). The probability of failure per hour associated with ASIL ratings is summarized in Table 6. Figure 12 shows an example breakdown of requirements on two redundant systems at ASIL B towards meeting an overall ASIL D target. Table 11 takes this a step further and shows what performance levels may be required of localization systems towards meeting overall system goals. This examines lateral localization requirements for lane determination and lane keeping for U.S. highway and local road geometries. Although centered around the ASIL D example, this table can also be used as a guide for the requirements outlined in Table 9 and Table 10. For example, if we consider a highway lane determination use case at ASIL B comprised of two redundant systems, each would need to be ASIL A outputting 0.41 m protection levels at $10^{-5}$ probability of failure per hour with a corresponding 95% accuracy of 0.18 m. This chart offers a quick look up for how such localization systems can be decomposed as a first step in overall system design.

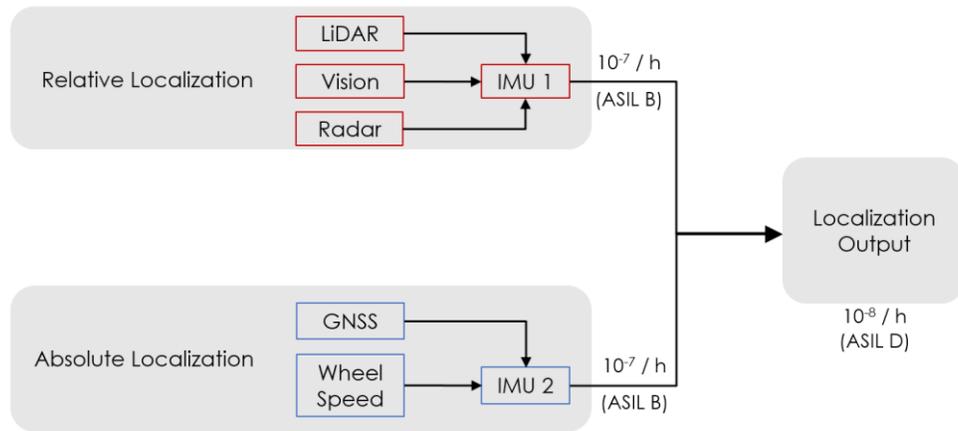

*Figure 12: Emerging localization system architecture in automated driving combining relative and absolute localization systems. This shows an example of system decomposition on requirements flow down for a desired ASIL D combined system composed of two independent ASIL B systems.*

*Table 11: Localization system decomposition analysis assuming a desired ASIL D level of performance. ASIL D systems can be decomposed into two independent ASIL B systems. Similarly, ASIL B systems can be broken into two ASIL A systems.*

| | | | Lane Keeping | | | | | | Lane Determination | | | | | |
|---|---|---|---|---|---|---|---|---|---|---|---|---|---|---|
| | | | ASIL D | | ASIL B | | ASIL A | | ASIL D | | ASIL B | | ASIL A | |
| | | | Sole System | | 2x Redundant | | 2+ Redundant | | Sole System | | 2x Redundant | | 2+ Redundant | |
| Percentiles [%] | Prob Failure / h | Gaussian Equiv. Num. σ | High way [m] | City [m] | High way [m] | City [m] | High way [m] | City [m] | High way [m] | City [m] | High way [m] | City [m] | High way [m] | City [m] |
| 99.999999 | 1.0E-08 | 5.73 | 0.41 | 0.21 | - | - | - | - | 1.11 | 0.89 | - | - | - | - |
| 99.99999 | 1.0E-07 | 5.32 | 0.38 | 0.19 | 0.41 | 0.21 | - | - | 1.03 | 0.83 | 1.11 | 0.89 | - | - |
| 99.9999 | 1.0E-06 | 4.89 | 0.35 | 0.18 | 0.38 | 0.19 | - | - | 0.95 | 0.76 | 1.02 | 0.82 | - | - |
| 99.999 | 1.0E-05 | 4.41 | 0.32 | 0.16 | 0.34 | 0.17 | 0.41 | 0.21 | 0.85 | 0.68 | 0.92 | 0.74 | 1.11 | 0.89 |
| 99.99 | 1.0E-04 | 3.89 | 0.28 | 0.14 | 0.30 | 0.15 | 0.36 | 0.19 | 0.75 | 0.60 | 0.81 | 0.65 | 0.98 | 0.79 |
| 99.7 | 2.7E-03 | 3.00 | 0.21 | 0.11 | 0.23 | 0.12 | 0.28 | 0.14 | 0.58 | 0.47 | 0.63 | 0.50 | 0.76 | 0.61 |
| 95 | 5.0E-02 | 1.96 | 0.14 | 0.07 | 0.15 | 0.08 | 0.18 | 0.09 | 0.38 | 0.30 | 0.41 | 0.33 | 0.49 | 0.40 |
| 68 | 3.2E-01 | 1.00 | 0.07 | 0.04 | 0.08 | 0.04 | 0.09 | 0.05 | 0.19 | 0.16 | 0.21 | 0.17 | 0.25 | 0.20 |

## CONCLUSION

Localization and mapping requirements for SAE Level 2+ ADAS semi-autonomous features represent near-term technology goals on the road to full autonomy. We presented localization and map feature georeferencing requirements in terms of accuracy and integrity based on vehicle dimensions, road geometry standards, and target level of safety. Due to the broad spectrum of semi-autonomous features, each requiring different levels of situational awareness, this was broken down by road determination, lane determination, and lane keeping. Integrity risk allocation leveraged techniques in civil aviation, drawing analogy between fully autonomous and semi-autonomous aircraft precision approach and landing systems. This allowed us to develop a risk model which includes the human-driver-in-the-loop. Based on this, integrity risk allocation is derived for mapping and localization at $10^{-6}$ probably of failure per hour of operation, similar in failure rate to ASIL B. Combining this with road geometry standards, requirements emerge for different road types and operation. For passenger vehicles operating on U.S. freeways, the result is a required lateral error bound of 2.26 m (0.90 m, 95%) for road determination, 1.11 m (0.44 m, 95%) for lane determination, and 0.41 m (0.16 m, 95%) for lane keeping. On local streets, the road geometry makes requirements more stringent, resulting in 1.93 m (0.77 m, 95%) for road determination, 0.89 m (0.36 m, 95%) for lane determination, and 0.21 m (0.08 m, 95%) for lane keeping. We selected to derive these requirements based on an equal sharing of the total error budget between the real-time localization solution and the map georeferencing, conceivably allowing fleet vehicles to act as mapping vehicles as well. Map georeferencing in a global common reference, such as the ITRF datum, is important as it enables data sharing and interoperability between autonomous platforms. This establishes situational awareness at the city level, providing yet another step towards closing the safety case in autonomy.

It should be emphasized that these requirements are not for one particular localization method or technology, but for the system of systems that will comprise it. In addition, the system must meet both 95% accuracy requirements and safety integrity level requirements in all weather and traffic conditions where operation is intended. Here, we present a framework for redundant system decomposition and localization requirements based on the limiting road geometry. These geometrical constraints represent the worst cases; with a-priori highly detailed maps of the environment, the road geometry will be known and hence localization resources can be adjusted on the fly to meet demand and could even be a semantic layer in the map itself. Achieving these requirements represents challenges in infrastructure, sensor, and algorithm development along with multi-modal sensor fusion to obtain the reliability levels needed for safe operation. Some techniques involving lidar, radar, and cameras rely on a-priori maps and give map-relative position. Others such as GNSS give global absolute position. Inertial systems maintain localization information when reliable updates are unavailable. Combining these and other technologies and selecting those most appropriate for the desired level of autonomous operation in a way that ensures integrity for safe operation is the challenge that lays ahead.

## ACKNOWLEDGMENTS

The authors would like to thank Xona Space Systems for supporting this work.

**BIOGRAPHIES**

**Tyler G. R. Reid** is a co-founder and CTO of Xona Space Systems whose focus is commercial satellite navigation services from Low Earth Orbit. Tyler previously worked as a Research Engineer at Ford Motor Company in localization and mapping for self-driving cars. He was also a Software Engineer at Google and a lecturer at Stanford University where co-taught the course on GPS. Tyler received his Ph.D. ('17) and M.Sc. ('12) in Aeronautics and Astronautics from Stanford where he worked in the GPS Research Lab and his B.Eng in Mechanical Engineering from McGill ('10). He is a recipient of the RTCA Jackson Award.

**Andrew Neish** is a co-founder and Director of Signals at Xona Space Systems where he works on signal design for next generation Low Earth Orbiting GNSS constellations. He completed B.Sc. ('14) in Mechanical and Aerospace Engineering from UC Davis and his M.Sc. ('17) in Aeronautics and Astronautics at Stanford University. He obtained his Ph.D. ('20) from the GPS Research Lab at Stanford with his thesis titled "Establishing Trust Through Authentication in Satellite Based Augmentation Systems." After graduating from Stanford, Andrew co-founded Xona Space Systems with an aim to bring modernized navigation capabilities to users worldwide.

**Brian Manning** is a co-founder and CEO of Xona Space Systems whose focus is commercial satellite navigation services from Low Earth Orbit. Brian previously worked at SpaceX as the Responsible Engineer for the Falcon 9 thrust structure, including obtaining human spaceflight ratings from NASA. He was an Engineering Manager at Power Solutions International working in high volume production alternative fuel automotive engines. Brian earned his MBA ('20) from the London School of Business, M.Sc. ('12) in Aeronautics and Astronautics from Stanford University, and B.S.E. ('10) in Mechanical Engineering from the Milwaukee School of Engineering.